\pdfoutput=1

\documentclass[11pt]{article}

\usepackage{acl}

\usepackage{times}
\usepackage{latexsym}

\usepackage{tabularx}
\usepackage{booktabs}
\usepackage{siunitx}

\usepackage{pgfplots}
\pgfplotsset{compat=1.17}
\usepgfplotslibrary{external, groupplots}
\usepackage{tikz}

\usepackage{subcaption}

\usepackage{hyperref}
\usepackage{listings}
\usepackage[T1]{fontenc}

\usepackage[utf8]{inputenc}

\usepackage{microtype}

\usepackage{inconsolata}

\usepackage{graphicx}

\usepackage{listings}
\usepackage{colortbl}
\usepackage{xcolor}
\usepackage{multicol}
\newcommand{\lowpercent}{\cellcolor[HTML]{dff7f3}}
\newcommand{\mediumpercent}{\cellcolor[HTML]{b6e3da}}
\newcommand{\highpercent}{\cellcolor[HTML]{92CEB8}}

\title{Low-resource Machine Translation: what for? who for? \\ An observational study on a dedicated Tetun language translation service}

\author{Raphael Merx$^1$, Adérito José Guterres Correia$^2$, Hanna Suominen$^3$, Ekaterina Vylomova$^1$\\
        $^1$School of Computing and Information Systems, The University of Melbourne\\
        $^2$Instituto Nacional de Linguística, Dili\\
        $^3$The Australian National University\\
        \texttt{rmerx@student.unimelb.edu.au}
        }

\begin{document}
\maketitle
\begin{abstract}


Low-resource machine translation (MT) presents a diversity of community needs and application challenges that remain poorly understood. To complement surveys and focus groups, which tend to rely on small samples of respondents, we propose an observational study on actual usage patterns of tetun.org, a specialized MT service for the Tetun language, which is the lingua franca in Timor-Leste. Our analysis of $100,000$ translation requests reveals patterns that challenge assumptions based on existing corpora. We find that users, many of them students on mobile devices, typically translate text from a high-resource language into Tetun across diverse domains including science, healthcare, and daily life. This contrasts sharply with available Tetun corpora, which are dominated by news articles covering government and social issues.
Our results suggest that MT systems for institutionalized minority languages like Tetun should prioritize accuracy on domains relevant to educational contexts, in the high-resource to low-resource direction.
More broadly, this study demonstrates how observational analysis can inform low-resource language technology development, by grounding research in practical community needs.

\end{abstract}

\section{Introduction}

While machine translation (MT) is often considered a single area for natural language processing (NLP), it covers a large range of applications, end-user requirements, and language directions. MT can be leveraged for everyday communication, specialized sectors (e.g., health or education), learning, or translation of official communication \cite{dew_development_2018, vieira_understanding_2021, herrera-espejel_use_2023, paterson2023machine, Merxetal2024}. It can be used by students in the classroom \cite{deng_systematic_2022}, workers in an office environment \cite{brynjolfsson2019does}, or by foreign travelers in public spaces \cite{carvalho2023attitudes}. Its reliability varies significantly depending on the size and domain coverage of available corpora \cite{dabre2020survey, khiu-etal-2024-predicting}, and depending on similarity between source and target languages \cite{xu2020boosting}.

\begin{figure}[t]
    \centering
    \includegraphics[width=0.5\textwidth]{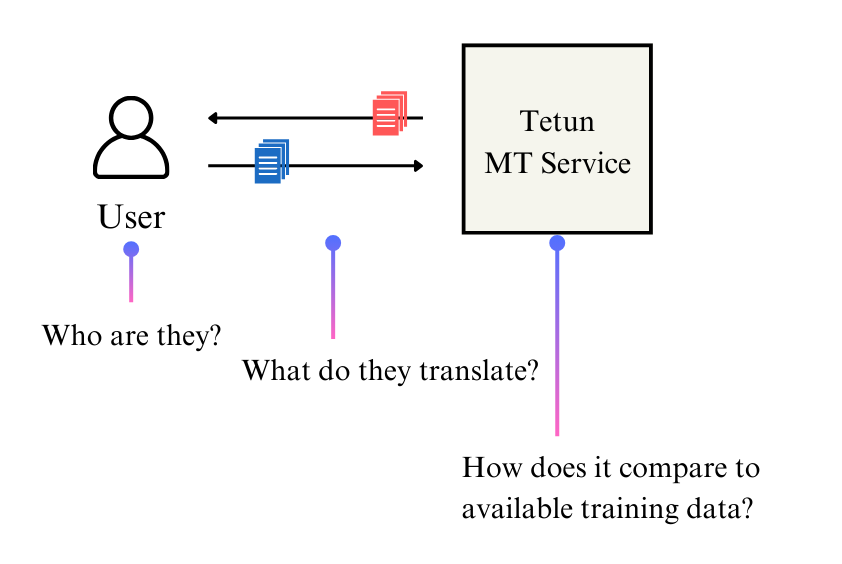}
    \caption{Overview of our approach and research~questions}
    \label{fig:real-vs-corpora}
\end{figure}

This disparity of potential applications and accuracy across  domains or language pairs is particularly salient when translating to or from low-resource languages \cite{haddow_survey_2022}. In low-resource scenarios, the combination of high variability both in MT accuracy across domains \cite{haddow_survey_2022} and of what native speakers value in NLP applications \cite{lent_what_2022}, has prompted NLP researchers to better anchor their work in local community perception, including through surveys, focus groups, and product design \cite{le_ferrand_fashioning_2022, mager_ethical_2023, blaschke_what_2024}. These approaches, however, could be complemented by observational studies that rely on inferring needs or lack thereof for MT across domains or demographics from end-user behaviors.

In this paper on low-resource MT, we work with the Tetun language,\footnote{Also called Tetun Dili, or Tetum.} which is the most spoken language in Timor-Leste \cite{Census2015}. We rely on a sample of server logs from \href{https://tetun.org}{tetun.org},
a volunteer-run MT service for the Tetun language, available in both website
and app\footnote{\href{https://play.google.com/store/apps/details?id=org.tetun.tetunorg}{play.google.com/store/apps/details?id=org.tetun.tetunorg}} format. Until Google Translate added Tetun support in June 2024, this service was the only widely available MT tool for Tetun. Tetun.org has over $70,000$ monthly active users, and translates over one million documents per month. From a sample of $100,000$ end-user translation text inputs, and their associated machine-translated output, collected between March and August 2024, we aim to respond to the following research questions (Figure~\ref{fig:real-vs-corpora}):
\begin{itemize}
    \item Who are the main users of tetun.org, what technology do they rely on, and what is the driver of their MT usage?
    \item What characterizes MT inputs, in terms of domain, length, direction of translation, and language pairs?
    \item How do MT inputs compare to available corpora for the Tetun language?
\end{itemize}


\paragraph{Tetun: low-resource, but institutional.} Following \citet{nigatu_zenos_2024}, we define why Tetun is low-resource in our view: (1) Tetun has very limited digital resources, with less than one million monolingual sentences available in open corpora \cite{de_jesus_labadain-30k_2024} (2) Tetun only has around one million native speakers \cite{Census2015} and (3) Tetun has very little NLP research coverage, apart from work by Gabriel de Jesus during his PhD candidature at the University of Porto \cite{de-jesus-nunes-2024-labadain-crawler, de_jesus_text_2024, de_jesus_exploring_2024}. However, Tetun is an institutional language, with both official status in Timor-Leste, and active usage in education and the media \cite{greksakova_tetun_2018}.  Our findings, while focused on Tetun, may extend to other institutionalized low-resource languages, but are less likely to apply to non-institutionalized or endangered ones.

\section{Related Work}

\paragraph{Community-driven approaches to NLP for low-resource languages.} Recent years have witnessed a growing focus on developing NLP technologies for low-resource and endangered languages. Researchers have identified that academic motivations can be at odds with the actual needs of language communities \cite{le_ferrand_fashioning_2022, liu_not_2022}, which made them advocate for a paradigm shift towards more community-centered approaches in NLP research \cite{lent_what_2022, bird_centering_2024}. Studies emphasize the importance of close collaboration with language communities, understanding their specific needs, and involving them in the design and development process of language technologies \cite{mager_ethical_2023, bird_centering_2024}. This approach acknowledges the heterogeneity among low-resource languages and their communities, recognizing that methodologies effective for one may not be universally applicable \cite{lent_what_2022, dolinska_akha_2024}.

\paragraph{Linguistic analysis through large-scale data gathering.} To our knowledge, this paper is the first large-scale analysis of real-world usage patterns in MT. It takes a similar approach to prior work leveraging large-scale secondary data in linguistics, including those predicting language learning difficulty \cite{schepens_big_2020}, understanding what language characteristics lead to a language being confused for another \cite{skirgard_why_2017}, and games that aim to incentivise the development of new language resources \cite{fiumara_nieuw_2022}.

\paragraph{Domain imbalance and low-resource MT}
Low-resource language corpora are often concentrated in specific domains, particularly religious texts \cite{marashian_priest_2025}, and lack the broad domain coverage typically found in their high-resource counterparts \cite{haddow_survey_2022}. As a consequence, NLP tools trained on these available corpora tend to lack domain robustness \cite{ranathunga_neural_2023}. This issue is particularly significant for MT systems, as demonstrated by \citet{khiu-etal-2024-predicting}, who find that domain similarity between training and test data is the largest driver of low-resource MT performance, more so than training data size or language similarity between source and target languages. Building on these findings, our analysis pays particular attention to differences of domain distribution between available corpora and actual translation requests.


\section{Methodology}
\label{sec:methodology}

We work with data from tetun.org, a specialized service for automated translation between Tetun Dili (\textit{tdt}) and any of English (\textit{eng}), Indonesian (\textit{ind}), or Portuguese (\textit{por}). The service also includes bilingual word dictionaries and a spell checker for the Tetun language, but we focus on the MT part.

The MT service uses a transformer model, and inference is run on the server side. The system was initially developed by fine-tuning a M2M100 multilingual model \cite{fan_beyond_2020} on a scraped parallel Tetun-English corpus. To optimize inference speed, this model was then distilled into a smaller transformer-base architecture. For Portuguese and Indonesian translation, the system uses English as a pivot language. The model's performance was evaluated using BLEU scores on a curated test dataset assembled by tetun.org volunteers.

We get access to the following data~sources:
\begin{enumerate}
    \item A sample of $100,000$ server logs from the service, all collected between March and August 2024. Each log contains data on timestamp, text input, MT output, source language code, requested target language code, and device operating system (OS).
    \item Summary monthly active users' data by country and device sourced from Google Analytics (for the website), Google Play Store,\footnote{\href{https://play.google.com/store/apps/details?id=org.tetun.tetunorg}{play.google.com/store/apps/details?id=org.tetun.tetunorg}} and Apple App Store.\footnote{\href{https://apps.apple.com/us/app/tetun-org/id1515892208}{apps.apple.com/us/app/tetun-org/id1515892208}}
\end{enumerate}

While data on the user base can be directly extracted from (2), domain analysis requires a choice of methodology that we describe next.

\subsection{Domain classification}

\paragraph{Components of domain. }
Recognizing that ``domain'' is often used interchangeably in NLP research to refer to topic (e.g., ``the health domain'') or source (e.g., ``the news domain''), we follow the methodology in \citet{saunders_domain_2022}, which defines domain as a combination of \textit{topic} (one or more subject matters covered by the text), \textit{provenance} (where the text comes from), and \textit{genre} (language syntax and style).

\subsubsection{Topic classification}


\paragraph{LDA optimization. }
We start by identifying topics using Latent Dirichlet Allocation (LDA, \citealp{blei_latent_2003}) on documents that are more than 5 words long, as very short documents do not offer enough context to identify their topic. We work with the Tetun side (either source Tetun text, or target machine-translated Tetun text), as this language is always present in any of the language pairs. The text is first lower-cased and punctuation is removed. We run LDA enhanced with bigrams that appear at least 5 times in the corpus for improved context. After running hyperparameter search on number of topics, passes, and alpha, we find a maximum topic coherence of 0.065 using 15 topics, 10 passes, and an asymmetric alpha. While this informs our choice of topic to categorize from, we qualitatively find that LDA is noisy for this task. For example, the health topic also includes words related to project management, such as ``objetivu'' or ``servisu''. It also performs poorly on the categorization of shorter documents, which are likely to contain words that are not part of any identified topics. However, we leverage these identified topics to formulate a list of topics to classify from.

\paragraph{Labels \& test set preparation. }
Through a manual review of each LDA topic, we land on the following list of topics to classify from:
\textit{Technology, Daily life \& personal experiences, Healthcare \& medicine, Law \& regulations, Business \& work \& employment, Government \& socio-economic issues, Education, Science \& research, Religion \& spirituality}. We map each of the 15 LDA topics to one of those topics, and our first author manually annotates a sample of 100 logs against this list of topics for evaluation, where each log can be classified against multiple topics.

\paragraph{Classification using pre-trained models. }
We evaluate other classification methods that use pre-trained models, either zero-shot transformer-based classifiers or LLMs. Given the lack of pre-trained models that cover Tetun, we rely on the high-resource side of the input-output pair, which can be in English, Indonesian or Portuguese. For zero-shot classifiers, we use the HuggingFace zero shot classification pipeline\footnote{\href{https://huggingface.co/tasks/zero-shot-classification}{huggingface.co/tasks/zero-shot-classification}} with the list of topics passed as candidate labels, in a multi-label mode. We evaluate \texttt{bart-large-mnli}\footnote{\label{bart-classifier}\href{https://huggingface.co/facebook/bart-large-mnli}{huggingface.co/facebook/bart-large-mnli}} and \texttt{deberta-v3-large-zeroshot-v2.0}\footnote{\label{deberta-classifier}\href{https://huggingface.co/MoritzLaurer/deberta-v3-large-zeroshot-v2.0}{huggingface.co/MoritzLaurer/deberta-v3-zeroshot-v2.0}} \cite{laurer_building_2023}. For LLMs, we assess LLama 3.1 8B \cite{dubey2024llama}, with a prompt that lists the domains to classify from, and that includes examples for in-context learning, shown in Appendix~\ref{sec:prompt-llama}.

\paragraph{Classification evaluation. }
We compare micro and weighted average scores on our test set for the different classification approaches in Table~\ref{tab:categorization_comparison}. We find that Llama 3.1 8B has the highest weighted averaged F1 score on both metrics for this classification task, followed by (in order) \texttt{deberta-v3-large-zeroshot-v2.0}, \texttt{bart-large-mnli}, and last LDA.

We therefore pick Llama 3.1 8B for topic classification, and run topic inference on the whole $100,000$ server logs, using a single Nvidia A100 GPU, which takes approximately 30 hours.  

\begin{table}[ht]
\centering
\small
\begin{tabularx}{\linewidth}{lSS}
\toprule
\textbf{Approach} & \textbf{Micro avg F1} & \textbf{Weighted avg F1} \\
\midrule
LDA & 0.21 & 0.19 \\
bart-mnli & 0.40 & 0.40 \\
deberta-nli & 0.66 & 0.68 \\
Llama 3.1 8B & 0.76 & 0.76 \\
\bottomrule
\end{tabularx}
\caption{Comparison of micro and weighted averaged F1 score for different domain classification approaches. LDA is run on the Tetun side, while other approaches are run on the high-resource side (English/Portuguese/Indonesian).}
\label{tab:categorization_comparison}
\end{table}

\subsubsection{Provenance classification}

After a manual review of the text samples used in topic classification, we propose the following categories for provenance: \textit{Education \& research material, News article \& press release, Organizational \& formal documents, Conversation \& correspondence, Literary fiction, Religious text, Other}. We use Llama 3.1 8B here as well, with a prompt that lists the sources to classify from, and that includes examples for in-context learning. On the same sample of 100 logs, which we manually annotate for provenance, we get a weighted F1 score of 0.73, which we deem satisfactory for classifying provenance, given the natural ambiguity of this task, especially for shorter documents. 

\subsubsection{Genre}

After recommendation from the Timor-Leste \textit{Instituto Nacional de Linguística} (INL, National Institute of Linguistics), under genre, we evaluate what fraction of Tetun MT inputs follow official spelling, as defined by the country's \textit{Matadalan Ortográfiku ba Tetun Nasionál} (Orthographic Guide for National Tetun, \citeyear{INL2002a}).

For calculating the percentage of words that follow INL spelling in each document, we apply the following preprocessing: (1) tokenize Tetun document using an open source Tetun tokenizer \cite{de-jesus-nunes-2024-labadain-crawler}, (2) ignore words that start with a capital letter as they are often proper nouns and (3) ignore tokens that start with a number or punctuation character. 
We further ignore inputs where less than 20\% of words were in the official spelling guide, as we empirically find that these documents are almost never in Tetun.

\subsubsection{Comparison with available monolingual corpora}

For each of the identified domain components (topic, provenance, genre), we provide a comparison between MT inputs and available Tetun corpora. For the latter, we select the Tetun ``clean'' split of the MADLAD-400 dataset \cite{MADLAD}, as it is the largest Tetun corpus available to date with $40,367$ documents. For classifying topic and provenance, given that we work with models that do not support Tetun, we first translate the MADLAD Tetun corpus to English using the MADLAD-3B model. Upon acceptance, we will release our translation of this corpus of $884,000$ Tetun sentences and their machine-translated text for future researchers to use. 

For all domain analysis, we weight inputs by their number of words. This is to ensure that very short MT inputs do not get as much weight as longer, more informative inputs, or that users who tend to translate several paragraphs at a time are not under-represented compared to users who translate shorter inputs.

\section{Results}

\subsection{User base}

\textbf{Country.} Most users are in Timor-Leste: 67,000 of 71,500 (93.4\%) Android monthly active users have their device registered in Timor-Leste, and 17,000 of 22,000 (77\%) website monthly active users are in Timor-Leste. These results support the hypothesis that the large majority of users are Timorese with English as a second language (ESL), as opposed to foreigners learning Tetun. While usage numbers can be noisy (for example, Timorese users who live abroad might have a mobile device registered outside of Timor-Leste, and foreigners in Timor-Leste using the website cannot be differentiated from local users), given the larger number of Timorese emigrating than people immigrating to Timor (INETL, \citeyear{timor-leste_national_institute_of_statistics_inetl_timor-leste_2022}), they are more likely to under-estimate than to over-estimate the proportion of Timorese users.

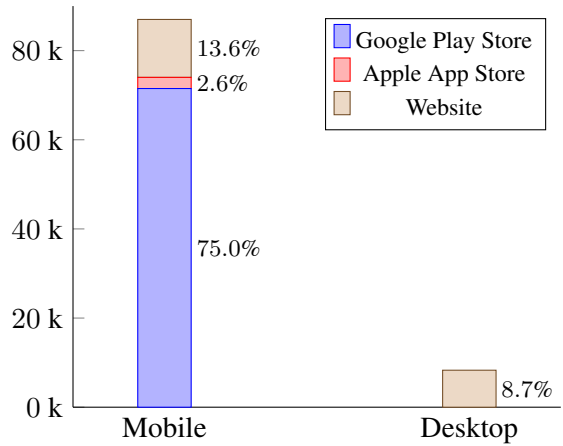
\begin{figure}
    \centering
    \begin{tikzpicture}
    \begin{axis}[
        ybar stacked,
        bar width=20pt,
        width=0.5\textwidth,
        enlarge x limits=0.3,
        symbolic x coords={Mobile, Desktop},
        xtick=data,
        ymin=0,
        ymax=90000,
        scaled y ticks=base 10:-3,
        ytick scale label code/.code={},
        yticklabel={\pgfmathprintnumber{\tick} k},
        legend pos=north east,
        axis x line*=bottom,
        axis y line*=left,
        legend style={font=\footnotesize},
        nodes near coords={
            \pgfkeys{/pgf/fpu=true}
            \pgfmathparse{\pgfplotspointmeta / 95300 * 100}
            \pgfmathprintnumber[fixed, precision=1, zerofill]{\pgfmathresult}\%
            \pgfkeys{/pgf/fpu=false}
        },
        nodes near coords align={right},
        nodes near coords style={font=\small, text=black, xshift=6pt},
    ]
    
    \addplot coordinates {(Mobile, 71500) (Desktop, 0)};
    
    \addplot coordinates {(Mobile, 2500) (Desktop, 0)};
    
    \addplot coordinates {(Mobile, 13000) (Desktop, 8300)};
    
    \legend{Google Play Store, Apple App Store, Website}
    \end{axis}
    \end{tikzpicture}
    \caption{Monthly active users by device and service. Note that users who use both the website and mobile app would be double counted.}
    \label{fig:mau-device}
\end{figure}

\textbf{Device.} Most users access tetun.org through their mobile device, either through the Android application ($71,500$ monthly active users), the iOS application ($2,500$ monthly active users), or the website in a mobile browser ($13,000$ monthly active users). Desktop users of the website represent $8,300
$ monthly active users, with an unknown number of these users being duplicate of mobile users. The distribution of users is illustrated in Figure~\ref{fig:mau-device}. We discuss the implications of this high representation of mobile devices, particularly Android devices, on NLP research, in Section~\ref{sec:recommendations}.

\begin{figure*}
    \centering
    \includegraphics[width=\linewidth]{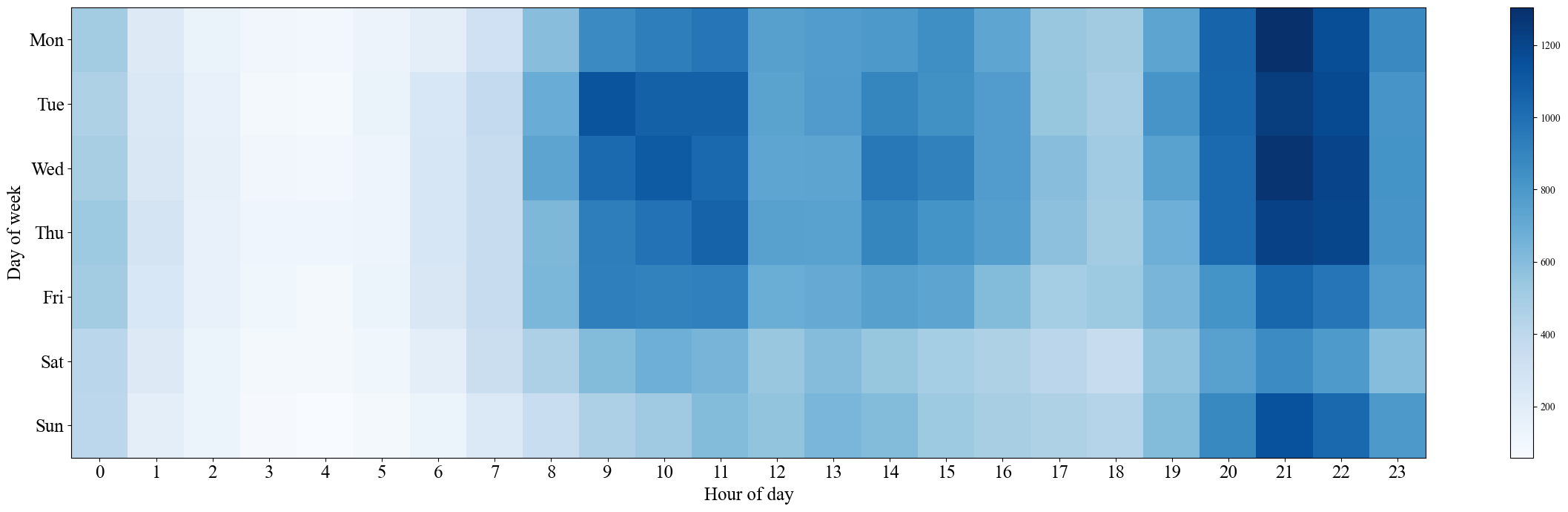}
    \caption{Number of requests by day / hour, in Asia/Dili (UTC+9). We observe a spike in the evenings before school days (Saturday is a school day in Timor-Leste), when students are more likely to prepare their homework.}
    \label{fig:hour-request-hist}
\end{figure*}

\textbf{Demographics.} Server usage patterns seem to indicate that most users are students who use tetun.org for educational purposes. This is supported by (1) the large proportion of requests that occur during evenings before school days (including on Sundays, see Figure~\ref{fig:hour-request-hist}), at times when office workers are less likely to use the service, and when students are more likely to do their homework; (2) the decrease in service usage during school holidays, with usage going back up the evening before school resumes; (3) domain analysis results presented in Section~\ref{sec:domain-coverage}.

\subsection{Translation direction and text length}
\label{sec:translation-direction}

\paragraph{Translation direction}
Using the source and target language codes present in each of the server logs, we extract summary information on the distribution by source and target language. We find that Tetun is more likely to be the target language (over 70\% of requests), with English to Tetun translation most in demand (46.7\% of requests), followed by Tetun to English (22.3\% of requests), Portuguese to Tetun (12.4\%), Indonesian to Tetun (11.4\%), Tetun to Portuguese (4.5\%), and Tetun to Indonesian (2.5\%) -- see Table~\ref{tab:lang-direction}. For all language pairs, the high-resource to low-resource direction is in higher demand than the low-resource to high-resource direction.

\begin{table}[t]
\centering
\small
\begin{tabularx}{\linewidth}{lc}
\toprule
\textbf{Direction}        & \textbf{Requests (\% of total)} \\
\midrule
English to Tetun                 & 46.7 \% \\
Portuguese to Tetun              & 12.4 \% \\
Indonesian to Tetun              & 11.4 \% \\
\textbf{Total Tetun as Target}                   & \textbf{70.6} \%       \\ 
\midrule
Tetun to English                 & 22.3 \% \\
Tetun to Portuguese              & 4.5 \% \\
Tetun to Indonesian              & 2.5 \% \\
\textbf{Total Tetun as Source}                   & \textbf{29.4} \%        \\ 
\bottomrule
\end{tabularx}
\caption{Distribution of translation requests by language pair}
\label{tab:lang-direction}
\end{table}

\begin{figure}[t]
    \centering
    \begin{tikzpicture}
        \begin{axis}[
            ybar,
            height=0.9\linewidth,
            width=\linewidth,
            bar width=9pt,
            xlabel={\small{Length of input text (number of words)}},
            xtick={0, 25, 50, 75, 97.5},
            xticklabels={0, 25, 50, 75, 95+},
            enlarge x limits=0.05,
            ymin=0,
            scaled y ticks=base 10:-3,
            ytick scale label code/.code={},
            yticklabel={\pgfmathprintnumber{\tick} k},
            xlabel near ticks,
            ylabel near ticks,
        ]
        \addplot coordinates {(2.5, 34175) (7.5, 17560) (12.5, 7052) (17.5, 4228) (22.5, 3199) (27.5, 2491) (32.5, 2379) (37.5, 2123) (42.5, 2019) (47.5, 1923) (52.5, 1713) (57.5, 1487) (62.5, 1271) (67.5, 1074) (72.5, 921) (77.5, 844) (82.5, 733) (87.5, 713) (92.5, 625) (97.5, 12143)};
        \end{axis}
    \end{tikzpicture}
    \caption{Histogram showing the frequency distribution of input text lengths measured by the number of words.}
    \label{fig:text-length-histogram}
\end{figure}
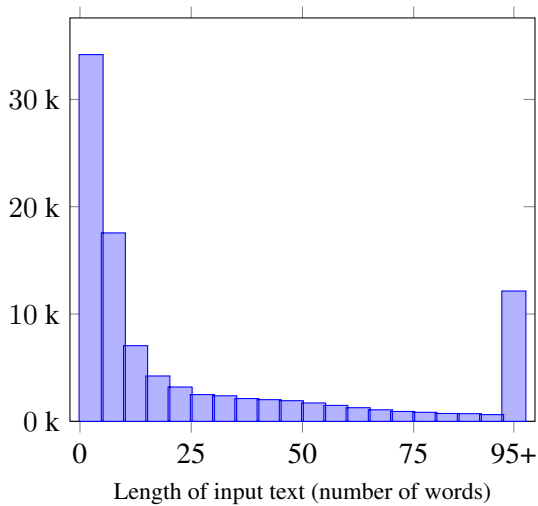

\paragraph{Length of translated text}
Leveraging an open source Tetun tokenizer \cite{de-jesus-nunes-2024-labadain-crawler} for Tetun, and the spaCy library\footnote{\href{https://spacy.io/}{https://spacy.io/}, version 3.8.2, model \texttt{xx\_ent\_wiki\_sm}} for English/Indonesian/Portuguese, we count the number of words in the input texts. We find that shorter text inputs are much more common, with a sharp decline in frequency as the text length increases. This does not include hits to the tetun.org dictionary, as the dictionary works offline. Overall, inputted text has a median of 8 words, with a median of 12 words when Tetun is the target of translation, and 5 words when Tetun is the source. We note however that this distribution has a long tail, with over 12\% of translations having 95 input words or more. A histogram of word count frequency is shown in Figure~\ref{fig:text-length-histogram}.

\subsection{Domain coverage}
\label{sec:domain-coverage}

In this section, we investigate what domains are most or least in demand by tetun.org users, and how they compare to domains present in available Tetun corpora, as represented by the MADLAD corpus. Given that domain similarity between training data and evaluation set is one of the main driver of MT accuracy \cite{khiu-etal-2024-predicting}, this investigation can inform future efforts on low-resource corpus gathering.

\subsubsection{Topic}
\label{sec:topic}

Using the input text and its associated MT-translated result from the server logs, we estimate what topics are most or least in-demand for text translation. Our results for topic coverage are shown in Table~\ref{tab:topic}. We include a comparison with topic coverage for the largest monolingual Tetun dataset currently available, MADLAD-400 \cite{MADLAD}.

We find overall that ``Science \& research'' is the topic most in demand, present in 34.2\% of MT inputs when weighted by word count, followed by ``Healthcare \& medicine'', ``Education'', and ``Business \& work \& employment'', all present in 20 to 25\% of inputs. Overall, ``Religion \& spirituality'' is the least frequent topic in MT inputs, which contrasts with the high representation of the religious domain in most low-resource corpora \cite{haddow_survey_2022}.

We observe a large discrepancy between user-translated text and available corpora, where the main topics covered by MADLAD (Government \& socio-economic issues, 83.1\%; Law \& regulations, 28.2\%) do not match with the main topics in demand by MT users.

\begin{table}[t]
\centering
\small
\begin{tabularx}{\linewidth}{lSS}
\toprule
\textbf{Topic} & \textbf{MT text} & \textbf{MADLAD} \\
\midrule
Science \& research & \mediumpercent 34.2\% & 3.2\% \\
Healthcare \& medicine & \mediumpercent 23.9\% & \lowpercent 11.9\% \\
Education & \mediumpercent 22.8\% & \lowpercent 11.3\% \\
Business \& work \& empl.. & \mediumpercent 21.2\% & 6.2\% \\
Daily life \& personal exp. & \lowpercent 17.2\% & 8.7\% \\
Government \& socio. issues & \lowpercent 16.2\% & \highpercent 83.1\% \\
Law \& regulations & 8.9\% & \mediumpercent 28.2\% \\
Technology & 8.1\% & 3.9\% \\
Religion \& spirituality & 6.8\% & \lowpercent 14.5\% \\
\bottomrule
\end{tabularx}
\caption{Topic distribution, where each document can cover several topics.}
\label{tab:topic}
\end{table}

We further analyze topic coverage depending on whether the source text is in Tetun or whether Tetun is the target language. When Tetun is the target, topics such as ``Science \& research'' (36.73\%), ``Healthcare \& medicine'' (25.52\%), and ``Education'' (23.14\%) dominate, reflecting user demand for translating high-resource content into Tetun, likely to support knowledge dissemination and accessibility. In contrast, when Tetun is the source language, translations are predominantly focused on ``Daily life \& personal experiences'' (48.64\%), indicating that users frequently translate everyday communication into high-resource languages. This highlights a divergence in translation use cases: translating into Tetun serves educational and professional purposes, whereas translating from Tetun often fulfills personal or practical communication needs. 

\subsubsection{Provenance}
\label{sec:provenance}

As illustrated in Table~\ref{tab:provenance_comparison}, we observe a large dominance of ``Education \& research material'', which alone represents more than half of MT inputs (55.9\%). ``Organizational \& formal documents'' comes as a distant second at 18\%. This again stands in contrast with monolingual corpora, where over 61\% of documents come from ``News articles \& press releases'', while only 7.5\% come from ``Education \& research material''.

Our findings on topic and provenance (Tables~~\ref{tab:topic} and \ref{tab:provenance_comparison}) are consistent with the hypothesis that most tetun.org users are students (who are more likely than the general population to translate educational material related to science, education or healthcare) and with \citet{de_jesus_labadain-30k_2024} who found that most Tetun text on the internet comes from news articles.

\begin{table}[t]
\centering
\small
\begin{tabularx}{\linewidth}{lSS}
\toprule
\textbf{Provenance} & \textbf{MT text} & \textbf{MADLAD} \\
\midrule
Education \& research material & \highpercent 55.9\% & 7.5\% \\
Organizational \& formal docs & \lowpercent 18.0\% & \lowpercent 17.3\% \\
News article \& press release & 10.2\% & \highpercent 61.2\% \\
Conversation \& correspond. & 8.7\% & 4.8\% \\
Literary fiction & 4.3\% & 0.9\% \\
Religious text & 2.5\% & 8.2\% \\
Other & 0.4\% & 0.3\% \\
\bottomrule
\end{tabularx}
\caption{Provenance distribution in MT and MADLAD.}
\label{tab:provenance_comparison}
\end{table}

\subsubsection{Genre}
\label{sec:genre}

\begin{figure}[t]
    \centering
    \begin{tikzpicture}
        \begin{groupplot}[
            /pgf/bar width=2pt,
            group style={
                group size=2 by 1,
                horizontal sep=0.7cm,
            },
            width=0.3\textwidth,
            height=4cm,
            ybar,
            ymin=0,
            scaled y ticks=base 10:-3,
            yticklabel=\empty,
            ytick=\empty,
            xlabel=\empty,
            xtick={25, 50, 75, 100},
            xticklabels={25\%, 50\%, 75\%, 100\%},
            xticklabel style={font=\small},
            enlarge x limits=0.05,
            xlabel near ticks,
            ylabel near ticks,
        ]
        \nextgroupplot[title={MT Tetun inputs}]
        \addplot coordinates {(23, 669) (27, 386) (31, 120) (35, 398) (39, 948) 
                             (43, 1691) (47, 3099) (51, 4726) (55, 8098) 
                             (59, 12395) (63, 19533) (67, 25622) (71, 27486) 
                             (75, 26565) (79, 25591) (83, 15145) (87, 16889) 
                             (91, 7773) (95, 4212) (99, 1975)};
        
        \nextgroupplot[title={MADLAD}]
        \addplot coordinates {(23, 129) (27, 221) (31, 5714) (35, 530) (38, 2007) 
                             (42, 6584) (46, 18367) (49, 52737) (53, 100213) 
                             (57, 307972) (61, 634412) (64, 1169741) (68, 1835142) 
                             (72, 2577109) (75, 3107333) (79, 3823647) 
                             (83, 4383662) (87, 3820195) (90, 1175277) (94, 83540)};
        \end{groupplot}
    \end{tikzpicture}
    \caption{Histograms of MT Tetun inputs and MADLAD Tetun texts by fraction of words in the official INL spelling guide.}
    \label{fig:inl-spelling-comparison-groupplot}
\end{figure}
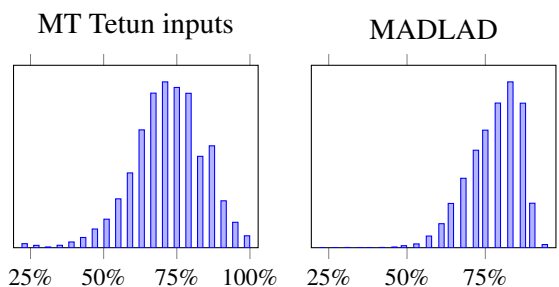

We observe that official (INL) spelling is only loosely followed in Tetun MT inputs, with a median of 71\% of words spelled correctly, compared to a median of 82\% in MADLAD. This is coherent with previous studies that note the high proportion of Tetun texts that do not follow the official spelling, including in the media \cite{greksakova_tetun_2018}. We also observe that variance in the rate of correct spelling is higher in MT inputs (standard deviation of 13\%) than in MADLAD (standard deviation of 8\%) -- see Figure~\ref{fig:inl-spelling-comparison-groupplot}.

\subsection{Effects of domain mismatch on LM modelling}

After uncovering a lack of similarity between MT inputs and available corpora in the previous section, we investigate downstream effects on language modelling, by comparing perplexity on MT Tetun inputs with perplexity on monolingual corpora using Goldfish Tetun\footnote{\href{https://huggingface.co/goldfish-models/tet_latn_full}{https://huggingface.co/goldfish-models/tet\_latn\_full}} \cite{chang_goldfish_2024}, a monolingual Tetun language model trained mostly (> 98\%) on MADLAD Tetun documents. We use the HuggingFace \texttt{evaluate}\footnote{\href{https://github.com/huggingface/evaluate}{https://github.com/huggingface/evaluate}, version 0.4.3} library for all perplexity measurements, and only measure on inputs of 5 words or more.

\paragraph{MT inputs vs monolingual corpora}
We compare perplexity on MT Tetun inputs with perplexity on the Labadain Tetun corpus \cite{de_jesus_labadain-30k_2024}, which is the second largest Tetun corpus after MADLAD, and was not used when training the Goldfish model. We also exclude Labadain sentences that are also present in MADLAD. We get an average perplexity of $1,774$ on MT Tetun inputs, compared to an average of $153$ for Labadain, which illustrates that MT inputs may contain vocabulary, structures, or topics that are underrepresented in scraped monolingual corpora, leading to lower language modelling performance and potentially increased translation errors.

\paragraph{Perplexity per topic}
Comparing average perplexity per topic, we find that documents in the ``Government \& socio-economic issues'' topic have a much lower perplexity (average of $632$) than documents from the ``Technology'', ``Business \& work \& employment'', or ``Daily life \& personal experiences'' domains ($3,434$, $1,871$ and $1,864$ respectively). This result illustrates that domain imbalance in monolingual corpora has downstream effect on language modelling ability for under-represented domains.

\paragraph{Perplexity per provenance}
Comparing average perplexity per provenance, we find that documents that come from ``News article \& press release'' have the lowest perplexity at $874$, compared to $1,320$ for ``Education \& research material'', $1,662$ for ``Organizational \& formal documents'' and $1,865$ for ``Conversation \& correspondence''. Here also, this result shows that document provenances that are under-represented in available corpora impact language modelling performance.

We report all perplexity results in Appendix~\ref{sec:perplexity-results}.

\section{Recommendations}
\label{sec:recommendations}

\paragraph{Anchoring MT evaluation in user demand.} Our results highlight the importance of: (1) broad domain coverage that includes education material, especially about science and health (2) accuracy when translating into, rather than from, the low-resource language, and (3) accuracy when translating very short documents (including individual words). This could start with an evaluation framework that finds a balance between these criteria. In particular, the sharp imbalance between the domain of available Tetun corpora and MT inputs leads us to recommend the systematic evaluation of MT accuracy in a wide variety of domains, and advise against measuring MT accuracy on domains that are over-represented in available corpora (e.g., news or religion), as this would risk giving a poor representation of end-user experience.

\paragraph{Meeting end-users where they are.}
\label{rec:hci}
We recommend a focus on meeting the needs of students who use MT on their mobile device. Given that students seem to represent a large portion of MT users, and that learning outcomes are an important social good, future study of MT perception and improvement for students in low-resource contexts would be beneficial, both to the research community and end users. Similarly, given the high cost and low reliability of internet connections in Timor-Leste \cite{asiafoundationDigitalYouth}, we recommend a heightened focus on MT inference on mobile devices, for example through further research into efficient MT, or through the creation of an open source framework for on-device MT that targets mobile devices.\footnote{Similar to the Bergamot / \href{https://browser.mt/}{Browser MT} project}

\section{Conclusion}

We present an analysis of MT usage for Tetun, a low-resource language, based on server logs from tetun.org, a widely-used MT service. Our study reveals key insights into user demographics, device preferences, translation directions, and domain coverage. We discover that a large share of users are likely students in Timor-Leste using mobile devices, translating education material from high-resource languages into Tetun. Our analysis also uncovers a significant mismatch between the domains of user-translated text and existing Tetun corpora.

Our recommendations include a stronger emphasis on domains that are not well covered by existing corpora (in particular health and science education material), accuracy for the high-resource $\rightarrow$ low-resource direction, performance on very short texts, and the development of mobile-optimized MT solutions. For human-computer interaction (HCI), we suggest focusing on MT perception and improvement for students in low-resource contexts.

This study provides the first observational analysis of real-world MT usage for a low-resource language. It contributes to a more nuanced understanding of the challenges and opportunities in this field. We hope these insights will guide future research efforts, ultimately leading to more effective and accessible MT solutions for speakers of low-resource languages.

\section*{Limitations}

\paragraph{System design decisions impacting numbers}
Design decisions made by the tetun.org volunteers might have impacted usage numbers. For example, apart from MT, tetun.org includes a multilingual word dictionary (with entry definitions and example sentences), and a Tetun spell checker, which might skew the number of users upward, with potentially some users using the app or website without using the MT functionality. Similarly, the authors chose to only allow translation between Tetun and three higher-resource languages (English, Portuguese, Indonesian), which raises the question of how adding support for more languages would impact usage and balance between languages.

\paragraph{Noise in reported data}
The methods used to determine user location may not accurately reflect the actual location and language background of the users. For instance, Timorese users living abroad might have devices registered outside Timor-Leste, and some users may access tetun.org through a VPN. These potential inaccuracies may affect our understanding of user demographics. Additionally, server logs only capture successful translation attempts, introducing potential bias: users may avoid tetun.org for domains or language pairs where it performs poorly, skewing the data toward areas where tetun.org is more effective.

\paragraph{Generalizability}
This paper focuses on Tetun, which might limit the applicability of our findings to other languages, given the variability of use cases, user needs and data availability within low-resource communities \cite{liu_not_2022}. While some of our findings are applicable to other low-resource languages, in particular those that are also institutionalized lingua franca, they are less likely to be applicable to much lower-resourced languages, in particular oral and endangered ones \cite{bird_centering_2024}.

\section*{Ethical considerations}

\paragraph{Data privacy}
While working with real end-user data poses privacy risks, we mitigate them through multiple strategies: (1) Server logs do not include user internet protocol (IP) addresses or personal account information; (2) To protect user confidentiality, we apply Named Entity Recognition using spaCy's \texttt{xx\_ent\_wiki\_sm} model to systematically remove names and addresses from translation inputs and outputs before analysis; (3) Our analysis exclusively reports aggregated, anonymized results, ensuring no individual user data can be traced or identified.


\paragraph{Ethics application and user consent}
We applied for and received a waiver of consent from our institution's ethics board, as obtaining explicit user consent was not possible given the anonymous nature of the service. This waiver was granted based on several factors: (1) users agree to research use of translation text under the service's privacy policy, which states: "We reserve the right to use translated text for research and product improvement purposes. All translated text is processed anonymously, ensuring that no personal identifiers are linked to the content"; (2) the service does not collect user accounts or identifying information; and (3) all analysis is conducted on aggregated data. Data management followed strict institutional protocols - all data was transferred directly from the service's server to our institution's high-performance computing environment, with no cross-border data flows, and access was restricted to authorized researchers only.


\bibliography{references, paper}

\appendix

\section{Prompt for domain classification}
\label{sec:prompt-llama}

\begin{lstlisting}[language=Python]
TEXT_PROMPT = """<text to classify>
{text}
</text to classify>"""

EXAMPLE_USER_1 = TEXT_PROMPT.format(text="Please meet me at the park at 3pm.")

EXAMPLE_RESPONSE_1 = """<rationale>
This text is about going to the park with a friend, a situation that is part of daily life experiences.
</rationale>

<domain>
daily life & personal experiences
</domain>"""

EXAMPLE_USER_2 = TEXT_PROMPT.format(text="Mindfulness-based interventions (MBIs) are increasingly being integrated into oncological treatment to mitigate psychological distress and promote emotional and physical well-being.")
EXAMPLE_RESPONSE_2 = """<rationale>
This text is about the use of mindfulness-based interventions in oncological treatment, a health topic. It seems to be from a scientific or research context.
</rationale>

<domain>
healthcare & medicine; science & research
</domain>"""

EXAMPLE_USER_3 = TEXT_PROMPT.format(text="The government has announced new policies to boost employment opportunities in rural areas.")
EXAMPLE_RESPONSE_3 = """<rationale>
This text discusses government policies aimed at improving employment, which relates to both governance and socio-economic issues, as well as business and employment.
</rationale>

<domain>
government & socio-economic issues; business & work & employment
</domain>"""

EXAMPLE_USER_4 = TEXT_PROMPT.format(text="Neraca adalah suatu laporan keuangan jadi di dalamnya pasti terdapat tiga bagian terpenting yaitu aset atau harta, liabilitas atau utang, dan ekuitas.")
EXAMPLE_RESPONSE_4 = """<rationale>
This text describes a financial report, specifically the balance sheet, which is an essential component of accounting and finance. It belongs to the business and finance domain.
</rationale>

<domain>
business & work & employment
</domain>"""

messages = [
    {"role": "system", "content": f"You are a linguist assistant, helping classify user-inputted text into a list of domains: <list of domains>{';'.join(TOPICS)}</list of domains>. You will be given text to classify, and asked the domains that this text belongs to. Only ever use the domains listed in <list of domains>"},
    {"role": "user", "content": EXAMPLE_USER_1},
    {"role": "system", "content": EXAMPLE_RESPONSE_1},
    {"role": "user", "content": EXAMPLE_USER_2},
    {"role": "system", "content": EXAMPLE_RESPONSE_2},
    {"role": "user", "content": EXAMPLE_USER_3},
    {"role": "system", "content": EXAMPLE_RESPONSE_3},
    {"role": "user", "content": EXAMPLE_USER_4},
    {"role": "system", "content": EXAMPLE_RESPONSE_4},
]
\end{lstlisting}

\newpage

\section{Perplexity results}
\label{sec:perplexity-results}

\begin{table}[h]
\centering
\small
\begin{tabular}{lr}
\toprule
\textbf{Topic} & \textbf{Average perplexity} \\
\midrule
Science \& research & $1,208$ \\
Healthcare \& medicine & $1,198$ \\
Education & $1,372$ \\
Business \& work \& empl. & $1,871$ \\
Daily life \& personal exp. & $1,864$ \\
Government \& socio. issues & $632$ \\
Law \& regulations & $1,274$ \\
Technology & $3,434$ \\
Religion \& spirituality & $1,127$ \\
\bottomrule
\end{tabular}
\caption{Perplexity per topic, as measured on Tetun MT input documents of 5 words or more.}
\label{tab:perplexity-topic}
\end{table}

\begin{table}[h]
\centering
\small
\begin{tabular}{lr}
\toprule
\textbf{Provenance} & \textbf{Average perplexity} \\
\midrule
Education \& research material & $1,321$ \\
Organizational \& formal docs & $1,662$ \\
News article \& press release & $874$ \\
Conversation \& correspond. & $1,866$ \\
Literary fiction & $2,119$ \\
Religious text & $1,473$ \\
Other & $3,876$ \\
\bottomrule
\end{tabular}
\caption{Perplexity per provenance, as measured on Tetun MT input documents of 5 words or more.}
\label{tab:perplexity-provenance}
\end{table}

\end{document}